# Movie Popularity Classification based on Inherent Movie Attributes using C4.5, PART and Correlation Coefficient

Khalid Ibnal Asad
Dept. of CS
AIUB
Dhaka, Bangladesh
khalidasad@aiub.edu

Tanvir Ahmed
Dept. of CS
AIUB
Dhaka, Bangladesh
tanvir@aiub.edu

Md. Saiedur Rahman
Dept. of CS
AIUB
Dhaka, Bangladesh
saied@aiub.edu

*Abstract* — Abundance of movie data across the internet makes it an obvious candidate for machine learning and knowledge discovery. But most researches are directed towards bi-polar classification of movie or generation of a movie recommendation system based on reviews given by viewers on various internet sites. Classification of movie popularity based solely on attributes of a movie i.e. actor, actress, director rating, language, country and budget etc. has been less highlighted due to large number of attributes that are associated with each movie and their differences in dimensions. In this paper, we propose classification scheme of pre-release movie popularity based on inherent attributes using C4.5 and PART classifier algorithm and define the relation between attributes of post release movies using correlation coefficient.

*Keywords- movie; IMDB; data mining; C4.5; PART; correlation coefficient*

## I. INTRODUCTION

Movie classification is a topic of interest both to academics and industry. Most of the classification schemes are focused towards user's preference on selecting future movies. But a classification scheme targeted for the future popularity of movie enables producers, financiers, academics or even viewers to understand the contributing factors that lead to movies success. This is because too many parameters of different degrees are related and finding a suitable way to represent all the information related to a movie in a single instance is a cumbersome task. Even if a way is found out to represent a movie the final choice of classifiers to generate the model requires considerable research. Again in case of post release movie the point of interest centers on the financial return. The problem of data representation and classification exits in this case also. So it is required to design an easily minable dataset along with appropriate classifiers that can be used to generate models to predict the classification of popularity of pre-release and post release movie.

The objective of this paper is firstly, to provide a suitable approach along with necessary factors that are to be considered for developing pre-release and post release movie datasets using Internet Movie Database (IMDB) data. Secondly, to select suitable classifiers based on dataset and target classification. Hence perform classification and evaluate the result. Lastly, find the attributes that contributes to the classification of movies, interpret those to provide better insight and in case the classification attempt fails find correlation coefficient amongst the attributes of the dataset.

## II. RELEVANT RESEARCH

The abundance of movie data in terms of review, rating or even detail information (for example the information maintained by IMDB) in the internet has encouraged many researches to formulate techniques to analyze the pattern in movie data. Most of the researches are devoted to develop recommendation systems of movie according to user reviews. In [1], research is directed towards classification of a movie based on the review written by a viewer. As each review is written focusing certain movie features the research extracts the feature-option pair and provides a summary of the movie in two broad categories: pros and cons. Research [2] was focused for KDD Cup 2007 and the task was to predict the probability that a user will give certain rating to a movie for a given list of 100,000 user–movie pairs. In [3], the research classified movie reviews into either positive or negative based on method consisting of two classifiers: SVMs and the scoring method.

User preferences are used in various online applications, such as movie recommendations sites, one-to-one marketing or now-a-days even in targeted advertisement. In most of the cases this data is drawn from the Netflix Prize dataset [4]. In [5] the research is directed towards the inclusion of user's context in addition to user's personality and recommendation vis-à-vis promotion. The research in [6] proposed constructing context aware and multi-applicable preference models using Bayesian networks to model user preferences. It develops a recommendation system that provides a list of movies to a user based on users certain inputs. This recommendation is done using Bayesian networks on previous data.

Besides this, [7] actually focused on the mining of original movie content as a whole that relates to this work of mining movie data. The concentration of the work was to focus on attributes relevant to the user ratings of movies, discovering the role of budget, significance of movies of a particular era and influence of particular actors or actresses in the rating of a movie. The thesis used Excel to cluster data into decade, a universal classifier query to see what factors are most relevant to the rating of a movie and finally generate a classifier that predicts movie rating based on actor, actress and directors involved in a movie. Although the preparation of pre-released movie dataset was motivated by few guidelines from the paper, the approach taken to mine the dataset and also the findings are totally different. Hence the approach as described in this paper is novel. The results obtained will be beneficial in two ways. Firstly, it will give a method to mine the unmanageable IMDB data. Secondly, it will enable anyone to understand the driving factors for movie popularity.

## III. PROCESS OVERVIEW

Basing on the objective we thought that it would be prudent to generate two separate datasets representing the pre-release and post release aspects of movies. Each attribute of datasets will then be evaluated in order to decide on the selection of classifiers so that optimum and less error prone result can be obtained. In any case if mining produces unsubstantial result then correlation coefficient between the attributes will be evaluated. Hence in order to meet the objective specified the entire process unfolds in three main steps, i.e. data collection, preparation and mining as seen in conventional machine learning approach. The entire process is elaborated in Fig.1

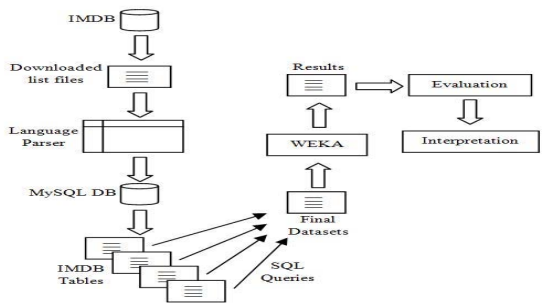

Figure 1. Process Architecture

## IV. DATA COLLECTION

### A. Obtaining Movie Data

The Internet Movie Database (IMDB) provides up to date movie information freely available on-line, across as many systems and platforms as possible. The database is updated weekly and is available over the internet. IMDB maintains the alternate interface site at [8] which describes various alternate ways to access the IMDB locally by holding copies of the data directly on local system. Local installations are available for a variety of formats. The files are compressed using GNU zip in order to save space and network bandwidth. Once uncompressed many of them take the format of LST (list file) and can be viewed using text editor/word processor. The database aims to capture any and all information associated with movies from any part of the world, starting with the earliest cinema to the very latest releases.

### B. Insight of IMDB Data

Once all the plain text data files are downloaded from [9], the next step is the local installation of data in a database. The database is logically organized in 4 broad categories with each containing a number of physical tables (list). The data is provided as forty-nine separate text files. The common factor linking the information in these files is the title of the movie, which is a title with the production year in brackets appended at the end to consider different versions of same movie, e.g. Godzilla (1954), Godzilla (1998). The files themselves are in a variety of formats with no conventions, such as Comma Separated Values (CSV), used. The data is laid out to be human readable not machine-readable. Much of the data is free text such as paragraphs giving film overviews or lists of quotations. This data is unsuitable for data mining without the additional use of natural language processing techniques for information retrieval or extraction [10].

### C. Using Third Party Tool for Database Generation

To minimize the effort spent on parsing all the text files and then converting each one to a table in a database, a third party tool name JMDB as obtained from [11] is used. It is an alternative way to navigate through movie information. JMDB automatically imports the list files and creates a MySQL database with tables and populates the tables with required data. JMDB constructed total 48 tables with each having movie id as the primary key and linked with the movies table. This provided the insight that raw IMDB data as obtained from [9] are unsuitable for data mining unless they are processed through some natural language processing tool. This limitation of IMDB data has not been known before until the data was downloaded. The limitation was mitigated using a third party free distributed tool (JMDB.

## V. DATA PREPARATION

### A. Factors Contributing to Movie Success

Before moving on to the data preparation for the mining purpose it is imperative to understand the success factor of a movie. The factors will help determine the necessary tables, generated in data collection phase, that are relevant to mining and eliminate the unnecessary ones. Factors that are thought to affect the commercial success may include [12]:
- Viewers.
- Star actor and actress.
- Market Trends.
- Budget.

Hence based on the factors described above, the variables responsible for the success of a movie are thought as follows:
- Country, year, language and movie genre.
- Director and cast ratings.
- Budget of film making, domestic, international and gross earnings.

Thus any pre-release movie must have country, year, language, director ratings, cast ratings and budget as the prime properties whereas all necessary financial information i.e. total domestic earning, total foreign earning, total worldwide earning will be contained in the post release movie information.

### B. Considerations Affecting Both Datasets

The first step required for the generation of datasets is to filter movies from TV shows, videos and mini-series as maintained by IMDB. The fact that each decade brings a psychological shift in terms of perception and views was also put into consideration. Hence movies released after year 2000 were only considered as they fell within the same decade which enjoyed the fruit of technology. The geographical consideration based on country and language also came to purview. Both the country and language was selected as USA due to more availability of data. Lastly movies receiving user votes less than 1000 were eliminated from the choice in order to reduce bias. At each step of the data generation process complex SQL queries were performed.

Before integration of all data into a single table it was required to insert classification information into the table. The data were classified based on user rating as provided by IMDB and in line with [7], shown in Table I.

Table I. CONSIDERATION FOR CLASS.

| Class | Rating |
|---|---|
| Excellent | 7.5 – 10 |
| Average | 5 – 7.4 |
| Poor | 2.5 – 4.9 |
| Terrible | 1 – 2.4 |

### C. Dataset 1 - Pre-Release Movie Dataset

The main transformation required was to calculate numerical ratings for the directors, actors and actresses. But being devoid of such rating, from [7] it was taken that a rating of a particular movie also reflects to some extent to the performance of directors and casts. The next step was to decide in what scale directors and casts should be represented. A certain movie consists of one, two or three directors. But the obvious fact is that in no case will a director with high average rank will tie up with a director with low standing. Thus averaging the total director rank based on [7] will be a logical step. Again movies consist of many casts and everyone contributes at some scale to the success of the movie. Therefore summing up the average rank of all the casts will be a logical step [7].

After the data generation it was found out that in the dataset average class dominated over the rest three while the terrible class consisted with only handful of instances. Hence it became imperative that the dataset should be adjusted so that no classes have dominance over the other. Thus the average class was filtered by taking 10 instances of each rank from 5.0 to 7.4 basing on user votes. This ultimately reduced the dataset to a more meaningful ratio as shown in Table II.

Table II. DISTRIBUTION OF DATA IN EACH CLASS.

| Class | Total no of instances |
|---|---|
| Excellent | 260 |
| Average | 250 |
| Poor | 288 |
| Terrible | 22 |

The final dataset is shown in table III.

Table III: DATASET 1 DESCRIPTION.

| Attribute Name | Attribute Description |
|---|---|
| id | IMDB movie id |
| Title | The original movie title |
| Year | Release year, year > 2000 and < 2011 |
| Language | Language = "English" |
| Country | Country = "USA" |
| Budget | Budget information in USD |
| Director Rank | Average rank of the directors |
| Male Cast Rank | Sum of ranks of all male casts |
| Female Cast Rank | Sum of ranks of all female casts |
| Votes | Original no of votes as given by IMDB users |
| Rating | IMDB users rating as generated by IMDB |
| Class | Classification of the movie based on rating |

### D. Dataset 2 – Post Release Movie Dataset

Total financial information of each movie was stored in plain text format in a column of budget table and linked to each movie with movie id. Hence it became a cumbersome task to parse the plain text instances in MySQL using the budget keywords as provided by IMDB. Therefore the obvious choice was to look for alternate approach. It was found out that Box Office Mojo a sister concern of IMDB contained all the financial information of movies. Thus an API was searched which can be called by passing the movie name and get the financial information. But Box Office Mojo provides only free data feeds [13] and for single movie at a time giving the gross earning. Hence further search continued and a PHP script from [14] was obtained that actually read the data from the Box Office Mojo site based on the input and provided all relevant financial information. The script file was modified a bit so that it could automatically insert the data into the required table.

The dataset generated consisted of 578 instances with four attributes namely; budget, domestic, foreign and worldwide all calculated in USD. The datasets were joined with the previous movie dataset and the total data distribution is as shown in table IV.

Table IV: DISTRIBUTION OF DATA IN EACH CLASS.

| Class | Total no of instances |
|---|---|
| Excellent | 172 |
| Average | 248 |
| Poor | 152 |
| Terrible | 6 |

The final dataset is shown in table V.

Table V: DATASET 2 DESCRIPTION.

| Attribute Name | Attribute Description |
|---|---|
| id | The movie id obtained from IMDB |
| Title | The original movie title |
| Budget | Budget information in USD |
| Domestic | Domestic earning in USD |
| Foreign | Foreign earning in USD |
| Worldwide | Worldwide earning in USD |
| Votes | Original no of votes given by IMDB users |
| Rating | IMDB users rating |
| Class | Classification of movie based on rating |

## VI. DATA MINING AND KNOWLEDGE DISCOVERY

### A. Selection of Classifiers

In order to mine IMDB data, a well-known data mining tool WEKA [16] was used. The consideration for choosing WEKA was driven by the dataset obtained from the previous section. Since the data has numeric data type with only the classification as nominal leading to the category of labeled data set [17]. Therefore it is needed to perform supervised data mining on the target data set [17] [18]. This narrowed down the choice of classifiers to only few, classifiers that can handle numeric data as well as give a classification (amongst a predefined set of classifications). Hence selecting C4.5 decision tree learning [19] [20] [21] and decision rules using PART [17] [18] became obvious. The attribute evaluation was also performed in order to find out the gain ratio and ranking of each attribute in the decision tree learning. In case for some data set data mining could not produce any suitable result then finding the correlation coefficient [22] was resorted to investigate if relation between attributes.

### B. Applying C4.5 Classifier on Dataset 1

The mining started with the C4.5 algorithm which in WEKA is implemented using J4.8 classifier [18]. The reduced error pruning [23] [24] was selected and the cross validation type was kept at default value of 10 folds since it generates a fairly accurate classification. After running, we found that WEKA generates the decision tree that can correctly classify 77.3562% instances and incorrectly classifies 22.6438%. The detailed breakdown of the classification accuracy is at Table VI.

Table VI. DETAIL CLASSIFICATION ACCURACY OF C4.5.

| Class | TP Rate | FP Rate | Precision | Recall |
|---|---|---|---|---|
| Excellent | 0.861 | 0.086 | 0.823 | 0.861 |
| Average | 0.69 | 0.162 | 0.65 | 0.69 |
| Poor | 0.774 | 0.083 | 0.835 | 0.774 |
| Terrible | 0.682 | 0.001 | 0.938 | 0.682 |

TP Rate and Recall [17] measures the amount of positive data that are really positive hence amongst 259 of the original excellent data the classifier after being trained and tested using 10 fold cross validation accurately classifies 86% of excellent data. But again it also misclassifies 8.6% of non-excellent data as excellent as provided by the FP Rate. On the other hand Precision gives the proportion of instances classified as positive that are really positive [17] and from the statistics above one can see that in case of excellent data that is 82.3%. The same amount of information for average, poor and terrible classes are found as well. From the above information it is well understood that the data given in precision column gives a definite performance view of the classifier. Therefore it can be said that C4.5 classifier achieved a 82.3% precision on classifying excellent class (high), 65% precision on classifying average class (moderate), 83.5% on classifying poor (high) and 93.8% on classifying terrible (very high). Now in order to visualize how many data has been misclassified in each class it is required to discuss the confusion matrix given in Table VII.

Table VII. CONFUSION MATRIX OF C4.5 CLASSIFIER.

| Class | Classified as | Excellent | Average | Poor | Terrible |
|---|---|---|---|---|---|
| Excellent | | 223 | 34 | 2 | 0 |
| Average | | 42 | 171 | 35 | 0 |
| Poor | | 6 | 58 | 223 | 1 |
| Terrible | | 0 | 0 | 7 | 15 |

The underlined numbers represent the original classification and rest misclassification. The decision tree as generated by the classifier is shown in Figure 2.

The decision tree uses the director rank as split on attribute and then the budget. The other attributes did not played any part in the forming of the pruned decision tree. This leads to the question of finding the gain ratio and the ranking of the attributes. To get the answers the attribute evaluator of WEKA was used. It was found out that director rank is ranked as the first having information gain 92.36% followed by budget having information gain 25.74%, male cast rank having information gain 15.53% and female cast rank having information gain 3.68%. Therefore it can be inferred from this tree that for data representation the decision tree depends upon the director average rank and the budget of the movie to classify it.

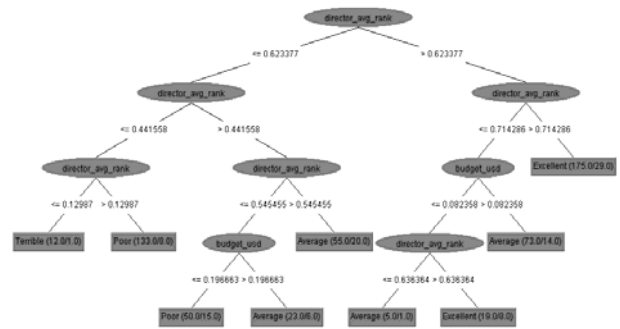

Figure 2. Decision Tree

### C. Applying PART Classifier on Dataset 1

In order to use the PART rule generator that uses C4.5 algorithm to generate partial decision trees and generate explicit rules [18] the rules subclass from the WEKA classify tab was selected followed by the PART classifier. Like decision tree, reduced error pruning and default 10 fold cross validation were selected allowing generalizing the rules in contrast specializing it. After running we found that WEKA generates the rules that can correctly classify 77.7234% instances and incorrectly classifies 22.2766%. Astonishingly this figure is the same as the C4.5 decision tree classifier as generated by the J4.8. In order to get a detailed breakdown of the classification accuracy we look at Table IX.

Table IX. DETAIL CLASSIFICATION ACCURACY OF PART.

| Class | TP Rate | FP Rate | Precision | Recall |
|---|---|---|---|---|
| Excellent | 0.826 | 0.063 | 0.859 | 0.826 |
| Average | 0.694 | 0.156 | 0.659 | 0.694 |
| Poor | 0.819 | 0.108 | 0.805 | 0.819 |
| Terrible | 0.591 | 0.001 | 0.929 | 0.591 |

Hence amongst 259 of the original excellent data the classifier after being trained and tested using 10 fold cross validation accurately classifies 82.6% of the excellent data as provided by TP Rate. But again it also misclassifies 6.3% of the non-excellent data as excellent provided by the FP Rate. On the other hand it is seen that in case of excellent data the precision is 85.9%. Same amount of information can be obtained for average, poor and terrible classes as well. As from previous discussion it is understood that data given in the precision column gives a definite performance view of the classifier. Hence it can be said that PART classifier achieved a 85.9% precision on classifying excellent class (high), 69.4% precision on classifying average class (moderate), 81.9% on classifying poor (high) and 92.9% on classifying terrible (very high). Now in order to visualize how many data has been misclassified in each class the confusion matrix has to be discussed as given in Table X.

Table X. CONFUSION MATRIX OF PART CLASSIFIER.

| Class | | Excellent | Average | Poor | Terrible |
|---|---|---|---|---|---|
| Excellent | Classified as | <u>214</u> | 41 | 4 | 0 |
| Average | | 32 | <u>172</u> | 44 | 0 |
| Poor | | 3 | 48 | <u>236</u> | 1 |
| Terrible | | 0 | 0 | 9 | <u>13</u> |

The underlined numbers represent the original classification and rest misclassification. Here also predominance of director average rank on generation of rules was observed. Some of the rules also used budget to along with director rank.

*D. Evaluating Information Gain of Dataset 2*

Before applying classifiers on the second dataset the information gain of each attribute towards the classification of the movie was found out using WEKA tool. The result of the Information Gain attribute selector is shown in Table XII.

Table XII. INFORMATION GAIN OF THE ATTRIBUTES.

| Attribute Name | Attribute Rank | Information Gain |
|---|---|---|
| foreign | 1 | 0.236 |
| worldwide | 2 | 0.206 |
| domestic | 3 | 0.191 |
| budget | 4 | 0.142 |

In this case a worst case scenario is visualized, where neither of the attribute has an information gain over 23%. Therefore it was certain that both C4.5 and PART classifier will fail to provide good result as both of them depends on the information gain to select an attribute to split on. The results are shown in Table XIII.

Table XII. DETAIL CLASSIFICATION ACCURACY OF C4.5.

| Class | TP Rate | FP Rate | Precision | Recall |
|---|---|---|---|---|
| Excellent | 0.209 | 0.086 | 0.507 | 0.209 |
| Average | 0.859 | 0.5 | 0.563 | 0.859 |
| Poor | 0.493 | 0.127 | 0.581 | 0.493 |
| Terrible | 0 | 0 | 0 | 0 |

The same level of accuracy was found in case of PART rule generator as shown in Table XIII.

Table XIII. DETAIL CLASSIFICATION ACCURACY OF PART.

| Class | TP Rate | FP Rate | Precision | Recall |
|---|---|---|---|---|
| Excellent | 0.174 | 0.064 | 0.536 | 0.174 |
| Average | 0.859 | 0.536 | 0.546 | 0.859 |
| Poor | 0.507 | 0.129 | 0.583 | 0.507 |
| Terrible | 0 | 0 | 0 | 0 |

*E. Finding Correlation Coefficient of Dataset 2*

The correlation is a way to measure how associated or related two variables are [22]. In order to find correlation the first attribute was chosen to be budget as it is obtained before the movie is released and the second attributes were domestic, foreign and worldwide respectively. This is shown in Table XIV.

Table XIV. CORRELATION COEFFICIENT.

| Attribute Names | Correlation Coefficient | Correlation |
|---|---|---|
| Budget + Domestic | 0.7 | Positive |
| Budget + Foreign | 0.76 | Positive |
| Budget + Worldwide | 0.76 | Positive |

Hence it is seen that a strong positive correlation exists between budget and rest of the attributes. To better understand the correlation a graph called scatter plot is used, where each point is represented by budget and its corresponding attribute (domestic, foreign and worldwide). The budget attribute is represented in the X axis and the corresponding other attributes were represented in the Y axis. A trend line was also used, which is best fit to the points and always goes in the direction of correlation i.e. in case of positive correlation from left to right and upwards. The scatter plots are shown in figures 3, 4 and 5.

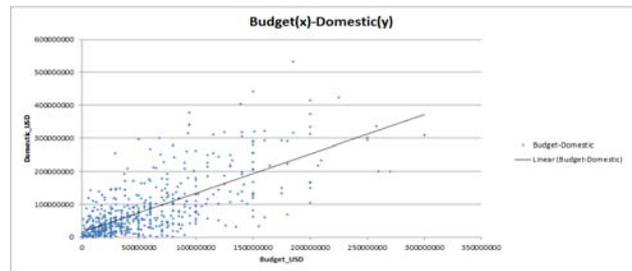

Figure 3. Scatter Plot of budget and domestic.

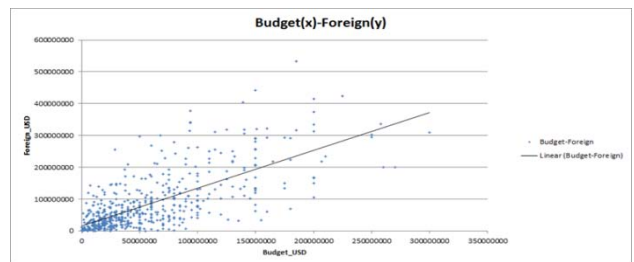

Figure 4. Scatter Plot of budget and foreign.

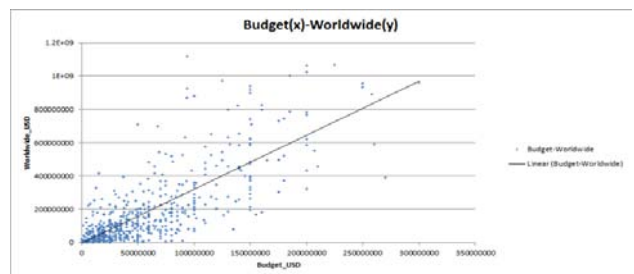

Figure 5. Scatter Plot of budget and worldwide.

The scatter plots and the correlation data together gave a very strong conclusion about the behavior of all the four attributes in the post release phase of the movie. The different approach to find any relation between budget of a movie and the other financial aspects didn't go awry. It was found out that budget is strongly related with other attributes namely; foreign earnings, domestic earnings and worldwide earnings. It was also seen that with the increase in budget the increase in amount in other three attributes is certain, which was observed from the trend line.

*F. Result Summary*

In case of the pre-released movie it was found that both classifiers astonishingly performed almost same. Both of them correctly classified 77% instances and misclassified 23% instances. It was also seen that average rank of directors played a vital role in attribute selection in both cases. The decision tree solely used director rank up to level 1 and then only used budget (in few cases) attribute in order to generate decision tree. The PART rule generator used director rank to generate rules, but also in few other cases used budget, male and female actor rank. Hence it can be concluded that higher director rank and larger budget contributed to better classification of movie and vice versa leads to lower classification.

In case of post release movie dataset correctly classified instances for both classifiers were only 56%, a very unstable result. Thus an alternate approach was needed, instead of trying to find out how each attribute contributed towards classification, find out correlation coefficient between budget and other three attributes in pair. Interestingly all correlation coefficients between budget and domestic, foreign and worldwide had a correlation coefficient of more than 0.7, typically a very high value. This showed how budget influenced post released financial aspects of movie. The trend line also showed a positive increase in all the scatter plots – another indication of strong correlation. Therefore it can be concluded that larger budget contributed to larger amount of return in turns of domestic earnings, foreign earnings and worldwide earnings of movie and vice versa.

VII. CONCLUSION

A machine learning approach has been provided for the prediction of movie popularity classification. This is a novel approach where the user rating decisions has been taken to purview along with inherent movie attributes to model the classification approach. An experimental insight has also been provided for the post release aspect of the movie that relates initial budget with each of the financial returns. The classification accuracy of both the decision tree and rules showed the dominance of director rank together with budget. But in case of post release movie dataset both classifiers failed to provide substantial result. Hence correlation coefficient between budget with each of foreign, domestic and worldwide was calculated and in each case a positive correlation was found which showed higher budget provided higher financial return and vice versa. The model and theoretical machine learning steps as shown in this paper will benefit various internet sites that are dealing with movie information. It will also aid producers and directors. It will also assist the film financing organizations to make decisions on movie rentals, streaming services, brand sponsorship, etc.